# A SELF-ADAPTIVE NETWORK FOR MULTIPLE SCLEROSIS LESION SEGMENTATION FROM MULTI-CONTRAST MRI WITH VARIOUS IMAGING PROTOCOLS

*Yushan Feng[1], Huitong Pan[1], Craig Meyer[1,2], and Xue Feng[1,2]*

[1]Springbok, Inc., Charlottesville, VA, USA
[2]Biomedical Engineering, University of Virginia, Charlottesville, VA, USA

## ABSTRACT

Deep neural networks (DNN) have shown promises in the lesion segmentation of multiple sclerosis (MS) from multi-contrast MRI including T1, T2, proton density (PD) and FLAIR sequences. However, one challenge in deploying such networks into clinical practice is the variability of imaging protocols, which often differ from the training dataset as certain MRI sequences may be unavailable or unusable. Therefore, trained networks need to adapt to practical situations when imaging protocols are different in deployment. In this paper, we propose a DNN-based MS lesion segmentation framework with a novel technique called sequence dropout which can adapt to various combinations of input MRI sequences during deployment and achieve the maximal possible performance from the given input. In addition, with this framework, we studied the quantitative impact of each MRI sequence on the MS lesion segmentation task without training separate networks. Experiments were performed using the IEEE ISBI 2015 Longitudinal MS Lesion Challenge dataset and our method is currently ranked 2$^{nd}$ with a Dice similarity coefficient of 0.684. Furthermore, we showed our network achieved the maximal possible performance when one sequence is unavailable during deployment by comparing with separate networks trained on the corresponding input MRI sequences. In particular, we discovered T1 and PD have minor impact on segmentation performance while FLAIR is the predominant sequence. Experiments with multiple missing sequences were also performed and showed the robustness of our network.

## 1. INTRODUCTION

Multiple sclerosis (MS) is an autoimmune disease of the central nervous system, in which inflammatory demyelination of axons causes focal lesions to occur in the brain, mostly in the white matter regions [1]. Magnetic resonance imaging (MRI) is a prevalent method to detect MS lesions, since MRI is very sensitive for detecting white matter lesions, especially with multi-contrast MRI containing T1, T2, proton density (PD) and FLAIR sequences [2]. In particular, T2 sequence reveals hyperintense lesions, T1 sequence reveals hypointense lesions, PD sequence is sensitive for detecting infratentorial lesions, and FLAIR sequence is sensitive for detecting supratentorial lesions [3].

Recently deep learning methods have been developed for MS lesion segmentation from multi-contrast MRI, including Recurrent Neural Networks (RNN) [4], Fully Convolutional Neural Networks (FCNN) [5] and Cascaded Convolutional Neural Networks (CNN) [6]. Notably, 3D U-Net [7], an architecture built upon the traditional CNN and specialized in image segmentation, is a commonly used DNN structure for lesion segmentation and other similar problems. Datasets for this task often contain multi-contrast MRI with several sequences. Since different MRI sequences reveal different information, they can be combined during the training and deployment of a network for the optimal performance through early-fusion, in which different sequences are concatenate as different input channels, or late-fusion, in which they are concatenated in a late stage of the network.

While combining all available sequences ensures the networks to exploit all information provided by the training data, it has a potential drawback when the trained networks are deployed into clinical practice. Imaging protocols often vary in different clinical sites, and it is unrealistic to expect the multi-contrast MRI obtained in all sites to match the availability and quality of every MRI sequence used during training of the networks. One straightforward solution to this problem is to train a network for each combination of multi-contrast MRI. However, it is impractical and error-prone due to the high computational cost of training DNN and the large number of possible combinations.

Therefore, it is of great practical significance to develop a MS lesion segmentation network that is able to adapt to various imaging protocols during deployment. Specifically, when certain MRI sequence used during training is unavailable during deployment, the segmentation network should achieve its maximal potential performance, i.e., the performance level of a separate network trained without this missing sequence. Furthermore, with this network, we can quantitatively evaluate the impact of each MRI sequence to the segmentation performance without training separate networks, thus providing meaningful insight for the optimization of imaging protocols.

We conducted our experiments using the IEEE ISBI 2015

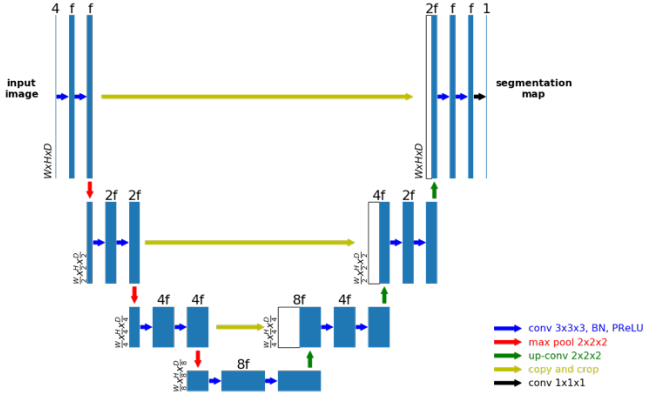

**Fig. 1.** Architecture of generic 3D-UNet

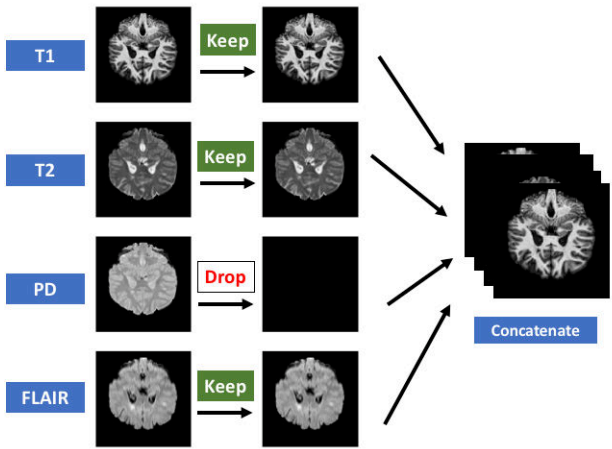

**Fig. 2.** Example of Sequence Dropout

MS lesion segmentation challenge (ISBI 2015) dataset [8], which contains 4 types of MRI sequence: T1, T2, PD, and FLAIR. Section 2 introduces our approach to build a segmentation network based on the optimized 3D U-Net architecture with non-uniform patch extraction and a novel technique called "sequence dropout" during training to ensure the 3D U-Net network trained on all 4 sequences can adapt to different imaging protocols with missing sequences. Section 3 details various experiments to assess our network's performance and robustness as well as the quantitative impact of each sequence to the segmentation performance. Discussions and conclusion are provided in Section 4.

## 2. METHODS

### 2.1 Optimized 3D U-Net with Non-uniform Patch Extraction

Our 3D U-Net implementation is illustrated in Fig. 1. The network takes in a 4D input formed by concatenating all 4 of the 3D MRI sequences together. The convolutional and deconvolutional paths of this 3D U-Net each contains 3 layers. The convolutional kernel size is 3×3×3 to handle 3D images. PReLu activation units are used for each layer. The number of root features is 96. The loss type is weighted cross entropy with a lesion/non-lesion ratio of 3 to 1.

As the original images are too large for the memory of a typical GPU, and they contain far fewer lesion voxels than non-lesion voxels [9], a non-uniform, patch-based strategy was used during training. The patch size was set as 64×64×64. When selecting training patches, we centered the patches at a voxel labelled as lesion by the ground truth mask with probability p = 0.99, and centered the patches at a non-lesion voxel with p = 0.01. Since lesion voxels usually cluster within a region, this strategy allows us to include sufficient lesion voxels in the training data. Combined with the weighted cross entropy loss function, it resolves the intrinsic unbalance between lesion and non-lesion voxels in the training data.

### 2.2 Sequence Dropout

Inspired by the dropout method widely used in training an artificial neural network, where hidden units in certain layers are randomly dropped to prevent the complex co-adaptations on training data and thus reduce overfitting [10], we developed a new technique called "sequence dropout" and randomly dropped certain MRI sequences when forming the training inputs. This technique ensures the network to learn the intrinsic information of each sequence and prevents it from learning the co-adaption of different sequences which is undesired for building a robust segmentation network.

The sequence dropout procedure is illustrated in Fig. 2 and implemented as follows: in the training stage, given the 4 MRI sequences from the training set, instead of immediately concatenating them together into a 4D input image to the network, we randomly decide to preserve $n$ ($0 < n < 4$) sequence(s) and substitute the other $4 - n$ sequence(s) with images containing zero value at each voxel. Then we concatenate the 4 MRI sequences together to form a 4D input image to the network. The rest of the network is unmodified.

All voxels in the input images are normalized to be from 0 to 1, and 0 represents the image background. By replacing the dropped-out sequence with an image filled by zeros, we essentially create a pure background with no additional information, thus preventing any unwanted meddling to the network. During deployment, for consistency, the missing sequence is also simulated by a background image with value zero at each voxel.

## 3. EXPERIMENTS

The ISBI 2015 dataset contains 5 training subjects and 14 testing subjects. Each subject has longitudinal MRI scans from 4 or 5 time points, and at each time point the subjects have 4 types of sequence: T1, T2, PD, and FLAIR.

In this dataset, two ground truth masks from two human experts are provided, and the network outputs will be evaluated on these two sets of ground truth masks. The Dice Similarity coefficients (DSC) calculated from the network outputs and each of the two ground truth masks are averaged as our main measurement of network performance. When generating labels during training, instead of choosing one specific ground truth mask, we merge the two masks and label voxels as 1 (lesion) or 0 (non-lesion) only where both masks agree; voxels where two masks contradict do not contribute to the network learning process and are labelled as 0.5. This strategy also modulates the inter-expert variability.

**Table 1.** DSC of the three methods over the validation set

| Methods | Missing Sequence | | | | |
|---|---|---|---|---|---|
| | N/A | T1 | T2 | FLAIR | PD |
| 4-seq (Generic) | **0.850** | 0.843 | 0.252 | 0.000 | 0.811 |
| Retrained 3-seq | N/A | **0.856** | 0.808 | 0.692 | 0.848 |
| 4-seq (Dropout) | 0.843 | 0.840 | **0.838** | **0.717** | 0.842 |

**Table 2.** DSC using sequence dropout

| Available Sequences | | | | |
|---|---|---|---|---|
| T1+PD | T2+FLAIR | T1+T2 | T1+FLAIR | FLAIR |
| 0.582 | 0.826 | 0.646 | 0.833 | 0.767 |

**Table 3.** ISBI Challenge Results

| Seq. Drop | DSC | Jaccard | PPV | TPR | LFPR | LTPR |
|---|---|---|---|---|---|---|
| No | 0.674 | 0.520 | **0.832** | 0.600 | **0.172** | 0.484 |
| Yes | **0.684** | **0.530** | 0.782 | **0.648** | 0.272 | **0.602** |

All networks used in the experiments were trained for 1000 epochs using the Adam optimizer with an initial learning rate of 0.0005. The training time was approximately 5 hours on a Nvidia TITAN V GPU.

### 3.1 3D U-Net on Validation Set

First, we trained a 3D U-Net using the longitudinal data from 4 subjects in the training set with all 4 available sequences using the conventional training strategy without sequence dropout. We evaluated this network on the validation set containing 4 time points of 1 subject with all sequences as well as under all scenarios where a certain sequence is unavailable and substituted by a zero-value background. In comparison, we trained 4 3D U-Net networks, each taking a unique set of 3 sequences as the training input, and evaluated the networks with their corresponding input sequences. Other network details were the same as in Section 2. We assume the results from the retrained networks with 3-sequence inputs as the maximal possible performance of our 3D U-Net design under 3 available sequences. Finally, we trained a 3D U-Net with all 4 sequences with sequence dropout and deployed it to all previous scenarios. Once we obtained the output segmentation masks from the network, we calculated the average DSC based on the ground truth masks. Table 1 shows the DSC achieved by the three approaches.

We observed that the network trained without sequence dropout achieved significantly lower DSC in scenarios without T2 and without FLAIR than its assumed maximal possible performance. It showed satisfactory results without T1 and without PD but completely collapsed without FLAIR (DSC: 0), whereas the 3-sequence network without FLAIR achieved reduced performance (DSC: 0.692). However, when sequence dropout was added to the network, its performance was similar as, or even slightly better than, the maximal possible performance achieved by the networks retrained with the corresponding input sequences.

In addition, we further evaluated the performance of this network when multiple sequences are missing in deployment. The DSC achieved by our network over a few scenarios are displayed in Table 2. Apparently, T2 and FLAIR are the more important sequences for MS lesion segmentation, while the absence of T1 and PD has minor impact on network performance. Even solely using FLAIR the network can still yield accurate segmentations with a DSC of 0.767.

### 3.3 3D U-Net on Test Set

Finally, we trained two 3D U-Net networks, one with sequence dropout and one without, using all 5 training subjects from the ISBI 2015 dataset. We submitted their outputs on the test set data to the challenge leaderboard. The performance metrics provided by the challenge are displayed in Table 3, where PPV denotes positive predictive value, TPR denotes voxel based sensitivity, LFPR denotes lesion based specificity and LTPR denotes lesion based sensitivity. The 3D U-Net without sequence dropout achieved good results and is currently ranked at 3[rd] in the ISBI 2015 challenge, whereas the network with sequence dropout achieves slightly higher overall score and is currently ranked at 2[nd]. The DSC of 0.684 achieved by the network with sequence dropout is currently the highest on the leaderboard.

To better illustrate the differences between the two 3D U-Nets with and without sequence dropout with one missing sequence during deployment, we selected a test case from the ISBI 2015 test set and visually compared the performance of the two networks as shown in Fig. 3. The network with sequence dropout showed more consistency in performance.

### 4. DISCUSSIONS AND CONCLUSION

In this study, we developed a self-adaptive network for MS lesion segmentation from multi-contrast MRI with various imaging protocol using the sequence dropout technique. This network can adapt to different combinations of input sequences and achieve the maximal possible performance similar as training a new network using the corresponding input sequences. On the other hand, the performance of the network trained without sequence dropout is not robust over various scenarios with certain missing sequence. Combined with non-uniform patch extraction, our 3D U-Net network has outstanding performance among all methods on the ISBI 2015 challenge leaderboard.

Similar as using unit dropout in the network layers, with sequence dropout in the network input, our network is able to learn the intrinsic information of each sequence instead of the co-adaption of different sequences. The fact that the network trained without sequence dropout completely collapsed without FLAIR means that the network learned to rely on FLAIR in all circumstances while other sequences are overshadowed. With sequence dropout, the network was forced to learn from any possible combinations of sequences to produce the segmentation output. It is also noteworthy that, despite the increased variability in training inputs when using sequence dropout, the training time did not need to be increased to achieve the same level of performance as before.

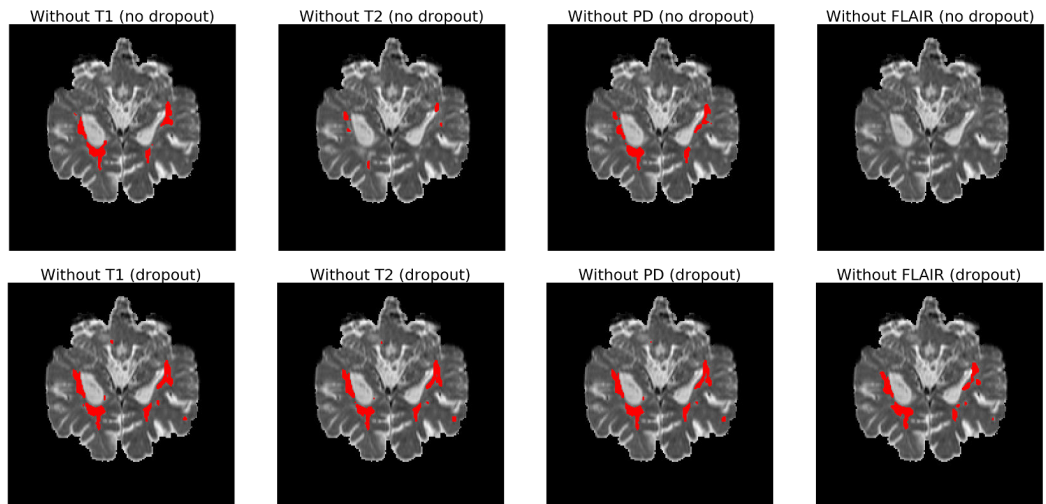

**Fig. 3.** Improvements from Sequence Dropout (Region in red represents predicted MS lesions)

For MS lesion segmentation, with this network, we were able to conveniently study the quantitative impact of different sequences without training separate networks. As expected, FLAIR appears to be the most important sequence in accurately delineating MS lesions since it has the same contrast as T2 but reduces the effect from fluid. T1 and PD are valuable but not indispensable for our network to generate solid results. In particular, T1 seems to contribute marginal information, as the network performance without T1 is largely unaffected. Further optimizations of MRI protocol can be explored and performed with these findings.

In addition, although many advanced networks have been proposed for this task, including DenseNet [9], we found that the 3D U-Net, with non-uniform patch extraction to address class imbalance and other optimizations, can still yield the best performance in practice.

While this study focused on the MS lesion segmentation task using 3D U-Net from multi-contrast MRI, the proposed sequence dropout can be easily transferred and applied to other network structures and applications, where the input sources may differ in deployment and the network needs to adapt to those various situations to achieve the maximal possible performance without retraining.